\documentclass[letterpaper]{article} 

\usepackage{aaai25}  
\usepackage{times}  
\usepackage{helvet}  
\usepackage{courier}  
\usepackage[hyphens]{url}  
\usepackage{graphicx} 
\usepackage{natbib}  
\usepackage{caption} 
\usepackage{algorithm}
\usepackage{algpseudocode} 
\usepackage{amsmath}
\usepackage{amssymb}
\usepackage{dsfont}
\usepackage{subfigure}
\newtheorem{pblm}{Problem}
\usepackage{newfloat}
\usepackage{listings}
\DeclareCaptionStyle{ruled}{labelfont=normalfont,labelsep=colon,strut=off} 
\floatstyle{ruled}
\newfloat{listing}{tb}{lst}{}
\floatname{listing}{Listing}
\title{Contextual Bandits for Maximizing Stimulated Word-of-Mouth Rewards}

\author {
    Ahmed Sayeed Faruk,
    Elena Zheleva
}
\affiliations {
    Department of Computer Science, University of Illinois at Chicago \\ Chicago, IL, USA\\
    \{afaruk2,
    ezheleva\}@uic.edu
}

\usepackage{bibentry}

\usepackage{comment}
\usepackage{commath}
\usepackage{algorithm}
\usepackage{algpseudocode}  
\usepackage{amsmath}
\usepackage[dvipsnames]{xcolor}
\usepackage{multirow}

\usepackage{graphicx}

\newcommand{\Bacc}{Recall@k}

\newcommand{\Lhse}{SpillCB}

\begin{document}

\maketitle

\begin{abstract}
Stimulated word-of-mouth is a strategy that promotes information sharing through prompts or incentives. Optimizing stimulated word-of-mouth through social networks requires identifying and targeting connected users who are most susceptible to spillover, a phenomenon where the influence of recommendations extends beyond the immediate audience to impact their connected users. The probability of spillover varies across individuals, and their connections, leading to heterogeneity. Understanding and accurately estimating the spillover probabilities among users in social networks is crucial for improving the effectiveness of stimulated word-of-mouth. To address this, we present a novel contextual multi-armed bandit framework that learns individual spillover probabilities and ranks connected users to maximize rewards from stimulated word-of-mouth. Experiments on real-world network datasets demonstrate that accounting for spillover heterogeneity enhances the targeting precision of top-$k$ connected users, boosting rewards and outperforming baseline methods that do not learn individual spillover effects.
\end{abstract}
\section{Introduction}
Stimulated word-of-mouth involves intentionally encouraging consumers to share information about a product or service, often using prompts or incentives provided by companies or marketers. Unlike organic word-of-mouth, which occurs naturally, stimulated word-of-mouth is facilitated through mechanisms such as rewards, discounts, or referral bonuses. Loyalty programs offering points for referrals exemplify this strategy. The proliferation of social networks has amplified the potential of stimulated word-of-mouth by simplifying the process of engaging and influencing consumers. For example, 
Alibaba integrates Sina Weibo (Chinese social network) into its e-commerce ecosystem, reflecting a trend where social platforms and marketplaces converge to enhance consumer influence and connectivity~\cite{zhao-mdpi-2022}.

Network spillover occurs when an action, such as a recommendation, on one user affects other users in a network, spreading information, behaviors, or attitudes. The impact of spillover is observed in numerous recommendation networks, such as social media, e-commerce, or content platforms~\cite{hollebeek-ir-2023, kuang-mis-2019, xu-tm-2022}. 
These spillover effects are heterogeneous, varying with users’ network positions, connections, and individual traits. For example, in a social network, sharing a recommendation can result in adoption by some friends and in lack of interest by others. 
Predicting and leveraging these heterogeneous spillover effects is crucial for optimizing stimulated word-of-mouth in networks. Here, we present preliminary work which leverages contextual multi-armed bandits (CMAB) for estimating such effects.


Contextual multi-armed bandits consider the scenario where decision-makers can observe the context of each possible recommendation (i.e., the arm) and infer the rewards of choosing a particular arm by observing the rewards of sequential recommendations. However, few works explore CMAB in networks where the recommendations can affect not only the direct object of recommendation but also its neighbors through spillover~\cite{faruk-arxiv-2023, iacob-sdm-2022,vaswani-icml-2017,wen-neurips-2017,wilder-aaai-2018}. While there is research that integrates CMAB with recommendation-dependent heterogeneous spillover in networks, it assumes that spillover probabilities are known~\cite{faruk-arxiv-2023}. In contrast, our focus is on learning these probabilities. 

The current literature on stimulated word-of-mouth overlooks the potential of leveraging the heterogeneous spillover effect, leading to sub-optimal performance and user satisfaction~\cite{gao-ieeetkde-2022}. To address this limitation, it is necessary to develop models that can learn and adapt to the varying spillover probabilities across different users in a network. This paper presents a novel framework to learn heterogeneous spillover probabilities in networks with the goal of recommending to each user the top-$k$ neighbors that are most likely to adopt the recommendation that the user themselves adopted. 
By advancing our understanding of these effects, we aim to enhance the performance of recommendation systems, delivering more effective and personalized user experiences while maximizing the rewards of stimulated word-of-mouth.


\section{Related Work}
Link weight prediction focuses on estimating the numerical value (e.g., strength, capacity) associated with an edge in a network~\cite{fu-ieeetkdd-2018}. Previous studies have investigated learning influence probabilities~\cite{goyal-wsdm-2010}. However, these concepts differ from the spillover probability considered in this work, which is tied to specific recommendations and spreads from the recommended user to its neighbor. In contrast, link weight or influence probabilities may exist independently of any recommendation.

Learning edge probabilities in networks has been studied using bandits framework for problems such as probabilistic maximum coverage and social influence maximization in viral marketing~\cite{chen-icml-2013}. However, most bandit research on networks~\cite{ iacob-sdm-2022,vaswani-icml-2017,wen-neurips-2017,wilder-aaai-2018} do not account for spillover effects. ~\citet{faruk-arxiv-2023} incorporate spillover effects in a bandit framework but assume that the spillover probabilities are given, without addressing the challenge of learning them.

While extensive research exists on top-$k$ recommendation problems~\cite{jamali-recsys-2009, song-icdm-2015, yang-recsys-2012}, limited attention has been given to identifying top-$k$ neighbors in social networks for stimulated word-of-mouth. \citet{aramayo-ms-2023} investigated the top-$k$ ad recommendation problem using a CMAB framework. However, their approach cannot be applied to the top-$k$ neighbor recommendation problem, as they assume a fixed number of possible ads, whereas users in networks have varying numbers of neighbors. A triad-based word-of-mouth recommendation model for socialized e-commerce focuses on recommending top-$k$ relevant products to a user for sharing with all neighbors~\cite{gao-ieeetkde-2022}. In contrast, our work focuses on recommending top-$k$ neighbors for sharing the same product.
\section{Problem Description}
\subsection{Data model}
We represent the data as a bidirected attributed network \( G = (V, E) \), where \( V = \{v_1, v_2, \ldots, v_n\} \) is the set of \( n \) nodes, and \( E = \{e_{ij} \mid 1 \leq i, j \leq n\} \) is the set of edges, with \( e_{ij} \) denoting an edge between nodes \( v_i \) and \( v_j \). The neighborhood of a node \( v_i \) is defined as \( \mathcal{N}_i = \{v_j \in V \mid e_{ij} \in E\} \), and each \( v_j \in \mathcal{N}_i \) is a neighbor of \( v_i \). Each node \( v_i \) is associated with a \( d \)-dimensional attribute vector, denoted by \( v_i.X \). Let \( v_i.y \in \{0, 1\} \) denote the activation status of node \( v_i \). Specifically, \( v_i.y = 1 \) indicates that \( v_i \) is activated, while \( v_i.y = 0 \) indicates no activation.

\subsection{Direct recommendation and spillover}
Direct recommendation refers to a system recommending a product directly to a targeted node in a network. A node activated due to such a recommendation is referred to as a system-activated node. Spillover occurs when a system-activated node impacts the activation of its neighbors by sharing the product. We assume that each user can only try to activate $k$ neighbors (e.g., due to the cost incurred by the user or the structure of the system-imposed sharing incentive) and the system's goal is to aid the user in picking the neighbors which are most likely to activate. Each edge \( e_{ij} \in E \) is associated with a spillover probability \( e_{ij}.p \in [0, 1] \), which represents the probability of activating node \( v_j \) due to spillover from system-activated node \( v_i \). The spillover probability can be asymmetric, i.e., \( e_{ij}.p \neq e_{ji}.p \), reflecting heterogeneity in spillover.

\begin{pblm}{\textbf{
Spillover probability prediction for top-$k$ neighbor recommendation}}
Given an attributed network \( G = (V, E) \), learn \(e_{ij}.p \) over \( T \) rounds such that the rewards associated with recommending the top-\( k \) neighbors of node \( v_i \) are maximized after \( T \) rounds.
\end{pblm}
For each system-activated node \( v_i \), the system's objective is to rank \( v_i \)'s neighbors by their spillover probabilities \( e_{ij}.p \), and select the top-\( k \) nodes to maximize the total rewards. Recommending the top-\( k \) neighbors based on the learned \( e_{ij}.p \) maximizes spillover rewards.

\section{Contextual bandits that learn individual spillover probabilities}
We consider a stochastic $2$-armed contextual bandit setup over \( T \) rounds. The set of arms is denoted as \( \mathcal{A} = \{a_0, a_1\} \). At each round \( i \in \{1, 2, \ldots, T\} \), a node \( v_i \in V \) arrives, and the learning agent observes the attributes of \( v_i \) and its neighbors \( v_j \in \mathcal{N}_i \). Without loss of generality, we only consider the nodes that the system was able to activate with its recommendations. The attribute vector \( e_{ij}.X\) is referred to as the context vector for edge \( e_{ij} \in E \). It includes $v_i.X$, $v_j.X$ and other relevant information, such as the product attributes. Each edge $e_{ij}$ is associated with two possible arms: $a_0$ and $a_1$, where $a_0$ corresponds to not recommending and $a_1$ corresponds to recommending the neighboring node $v_j$ to node $v_i$. We denote the assigned arm with $e_{ij}.t$ for edge $e_{ij}$. A reward is generated for recommending a neighboring node, corresponding to the activation of an inactive neighbor \( v_j \) due to spillover from the system-activated node \( v_i \). For example, a reward of 1 is assigned to the system-activated node if a neighboring node is activated due to spillover; otherwise, the reward is 0. The rewards are cumulative when multiple neighbors are activated. We assume that the reward for an action $e_{ij}.a$ on edge $e_{ij}$ is a function of the context vector $e_{ij}.X$ and the action that needs to be learned:
$R(e_{ij}.X, e_{ij}.a)  = F (e_{ij}.X, e_{ij}.a) + \xi_{ij}$, where $\xi_{ij}$ is zero-mean noise. The system recommends the top-$k$ neighbors based on the highest predicted rewards of \( a_1 \) arm, \( R(e_{ij}.X, a_1) \). The estimated spillover probability corresponds to the probability that the reward \( R(e_{ij}.X, a_1) \) coming from neighbor $v_j$ is $1$. The predicted set of top-$k$ neighbors is defined as $N_i \leftarrow \underset{v_j}{\arg top_k} R(e_{ij}.X, a_{1})$. We define the optimal set of top-$k$ neighbors as
$N^{*}_{i} \leftarrow \underset{v_j}{\arg top_k} R(e_{ij}^*.X, a_{1})$, where $e_{ij}^*$s yield maximum expected rewards for neighbor $v_j$ at round $i$. Therefore, the cumulative regret of bandit learning over $T$ rounds is formulated as: $Regret \leftarrow \sum_{i = 1}^{T} (\sum_{a \in N^{*}_i}R(e_{ia}.X, a_{1}) - \sum_{b \in N_i}R(e_{ib}.X, a_{1}))$. The objective of the contextual bandit is to learn a strategy for selecting the top-$k$ neighbors such that the expected regret is minimized. 

The pseudo-code of our framework is included in Algorithm $1$. The framework leverages CMAB algorithms to learn spillover probability for top-$k$ neighbor recommendations which operate in two distinct phases: exploration and exploitation. The exploration phase aims to diversify edge selections, which lasts for the initial $z\%$ of the total $T$ rounds. In this phase, $k$ neighbors are randomly selected and recommended to the system-activated node. After the exploration phase, the algorithm transitions to the exploitation phase. Here, the top-$k$ neighbors are recommended based on the descending order of their estimated reward for arm $a_1$, as predicted by the CMAB algorithm.  
\section{Experiments}
In this section, we evaluate the performance of $\Lhse$ on real-world network datasets.
\subsection{Data representation}
\subsubsection{Real-world datasets.} 
We consider two real-world attributed network datasets: Flickr\footnote{\label{note1}All datasets available at https://renchi.ac.cn/datasets/} and Facebook\footref{note1}. We consider the concatenation of $v_i.X$ and $v_j.X$ as the attribute vector for each edge \( e_{ij} \in E\). The spillover probability $e_{ij}.p$ for each edge $e_{ij} \in E$ is synthetically generated as a weighted function of cosine similarity(CS), level similarity(LS),  topological overlap(TO), diffusion importance(DI), and degree product(DP), following \cite{qian-nature-2017}:
\begin{multline*}
   e_{ij}.p \leftarrow f (w_0 + w_1 * CS (v_i, v_j) + w_2 * LS(v_i, v_j) + \\ w_3 * TO(v_i, v_j)  + w_4 * DI(v_i, v_j) +
    w_5 * DP(v_i, v_j))
\end{multline*}
where $w \sim \mathcal{N}(5,2)$  are the weights sampled from a normal distribution,
and $f$ is a min-max scaling function for normalization.

\begin{algorithm}[tb]
	\caption{Learning heterogeneous spillover to maximize word-of-mouth rewards ($\Lhse$)}
	\begin{algorithmic}[1]
	    \State \textbf{Input:} Number of rounds $T$, off-the-shelf CMAB (e.g., LinUCB) hyperparameters, exploration ratio $z$, $k$ 
       \State \textbf{Output:} Top-$k$ neighbor recommendations for each node $v_i$ that gets activated by the system in each round $i$
	    \For {each arm $a \in \mathcal{A}$}
	        \State Initialize arm parameters
	    \EndFor
		\For {each round $i \in \{1,2,3,\ldots,T\}$}
    		\State A node $v_i$ arrives with context vectors for edges $e_{ij}$, $\{e_{ij, a}.X\}_{a \in \mathcal{A}}$ and gets activated by the system 
		    \State Estimate expected rewards of each edge $e_{ij}$ for $a_1$, $E[R(e_{ij}.X, a_1)]$ using off-the-shelf CMAB algorithm
            \If {$i \leq z * T$}  \Comment{\textit{Exploration phase}}
    	   \State Find $N_i = \underset{v_j}{\arg random_k} E[R(e_{ij}.X, a_1)]$
            \Else \Comment{\textit{Exploitation phase}}
    	   \State Find $N_i = \underset{v_j}{\arg top_k} E[R(e_{ij}.X, a_1)]$
           \EndIf
                \For {$v_j \in N_i$}
                    \State $e_{ij}.t \leftarrow a_1$ 
                    \State Record $R(e_{ij}.X, a_1)$; update parameters for ${a}_1$
                \EndFor 
		\EndFor 
	\end{algorithmic} 
\end{algorithm}
\begin{figure*}[t]
    \centering
     \begin{subfigure}
         \centering \includegraphics[width=0.24\textwidth]{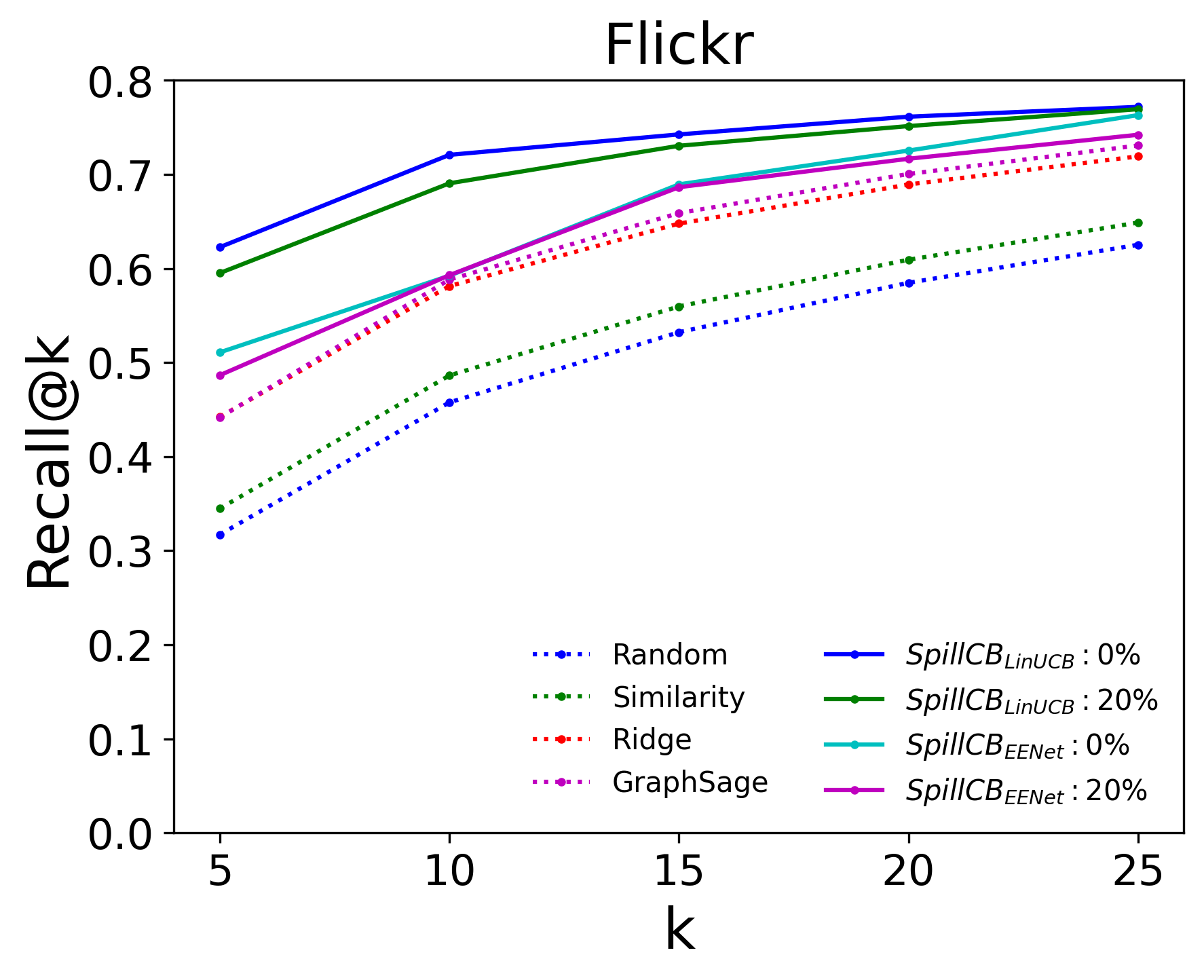}
     \end{subfigure}
     \begin{subfigure}
         \centering \includegraphics[width=0.24\textwidth]{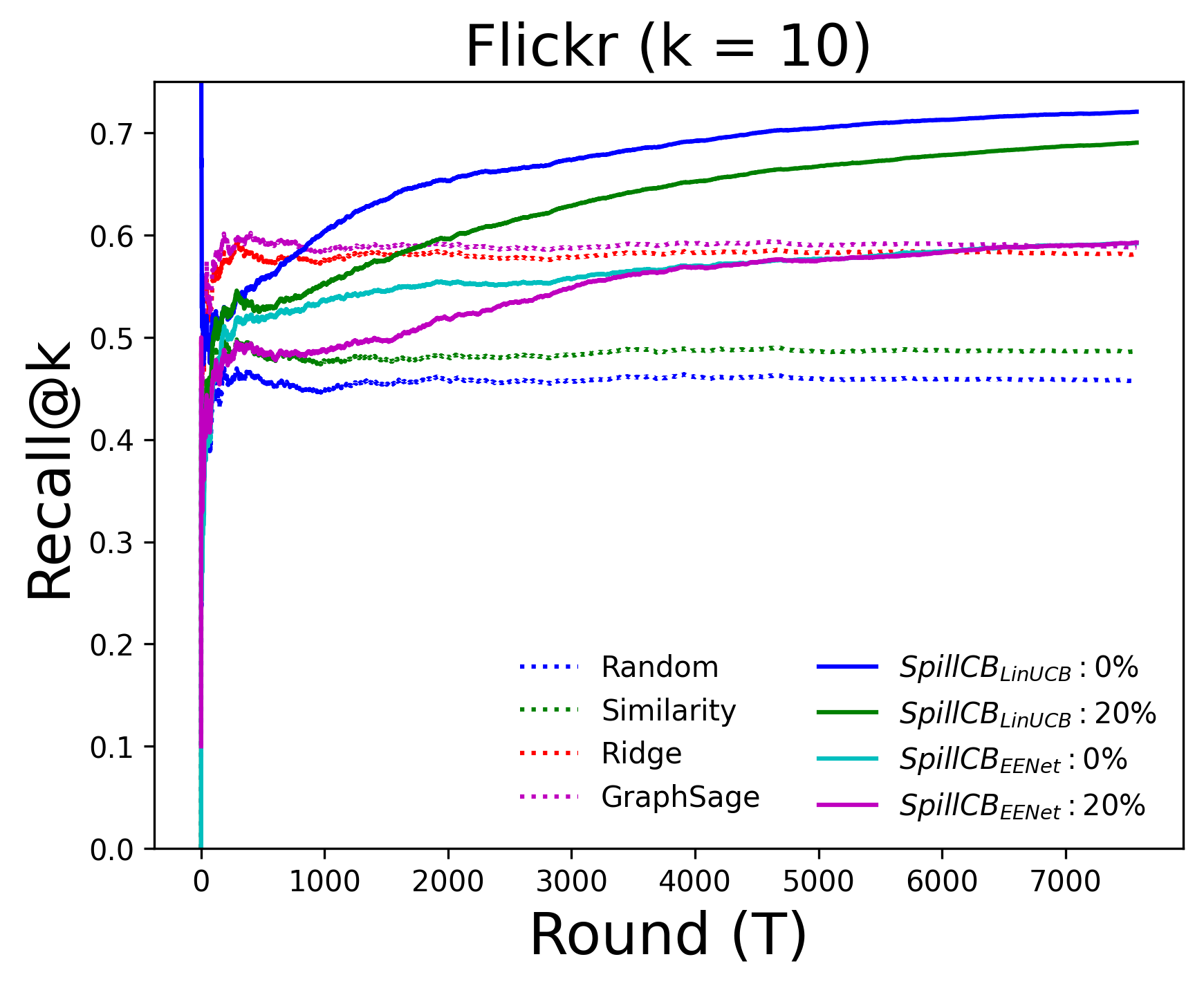}
     \end{subfigure}
     \begin{subfigure}
         \centering \includegraphics[width=0.24\textwidth]{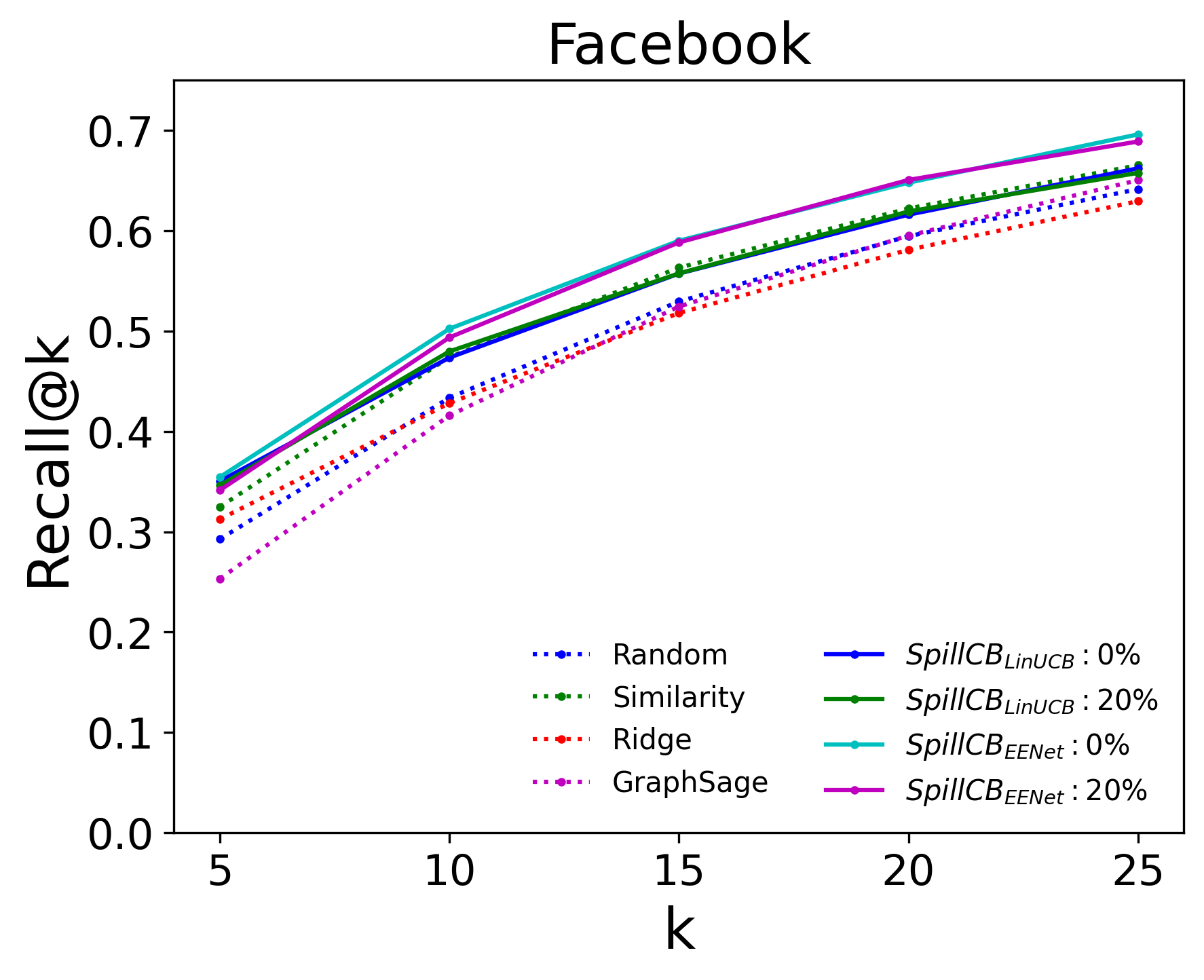}
     \end{subfigure}
     \begin{subfigure}
         \centering \includegraphics[width=0.24\textwidth]{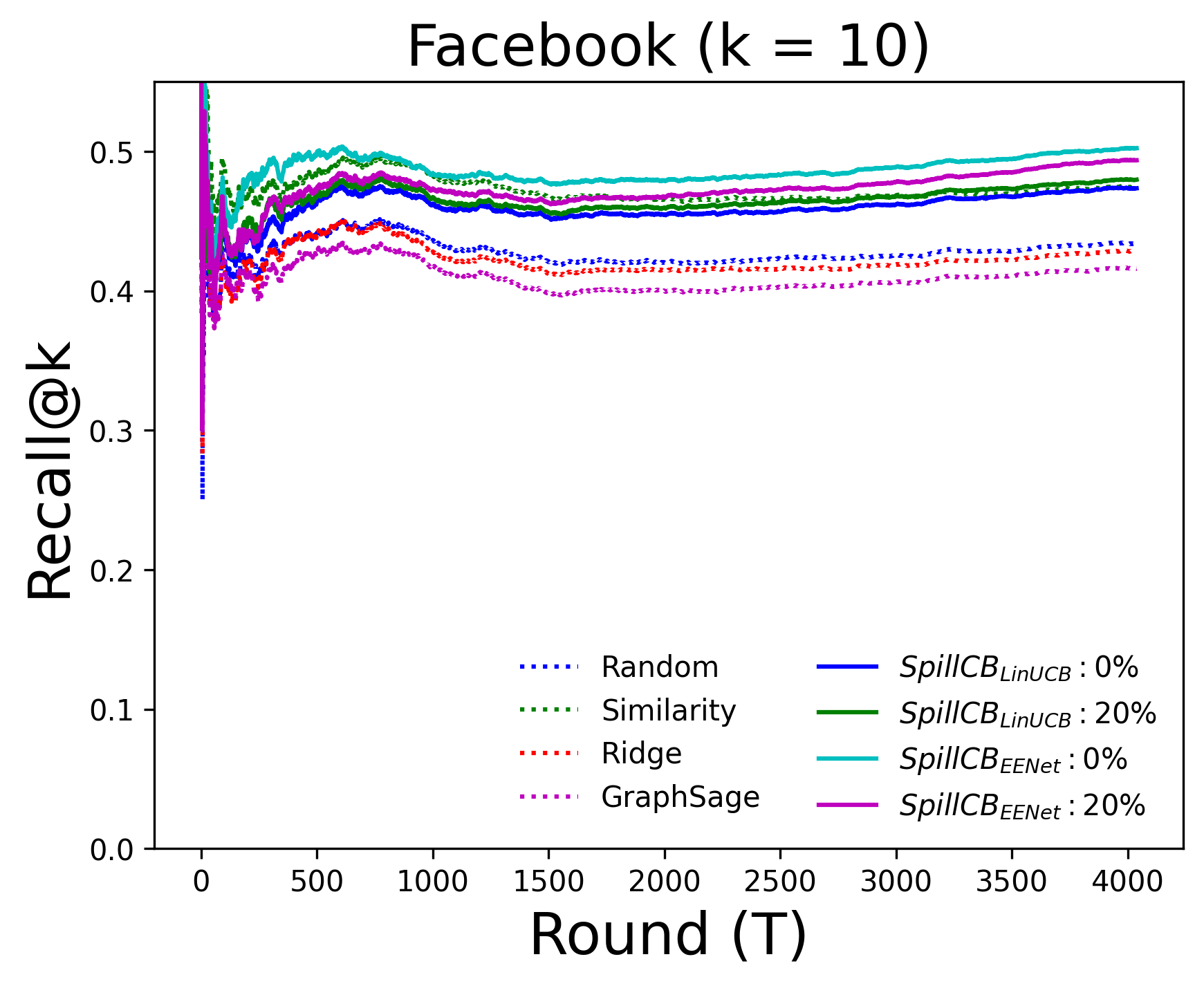}
     \end{subfigure}
     \begin{subfigure}
         \centering \includegraphics[width=0.24\textwidth]{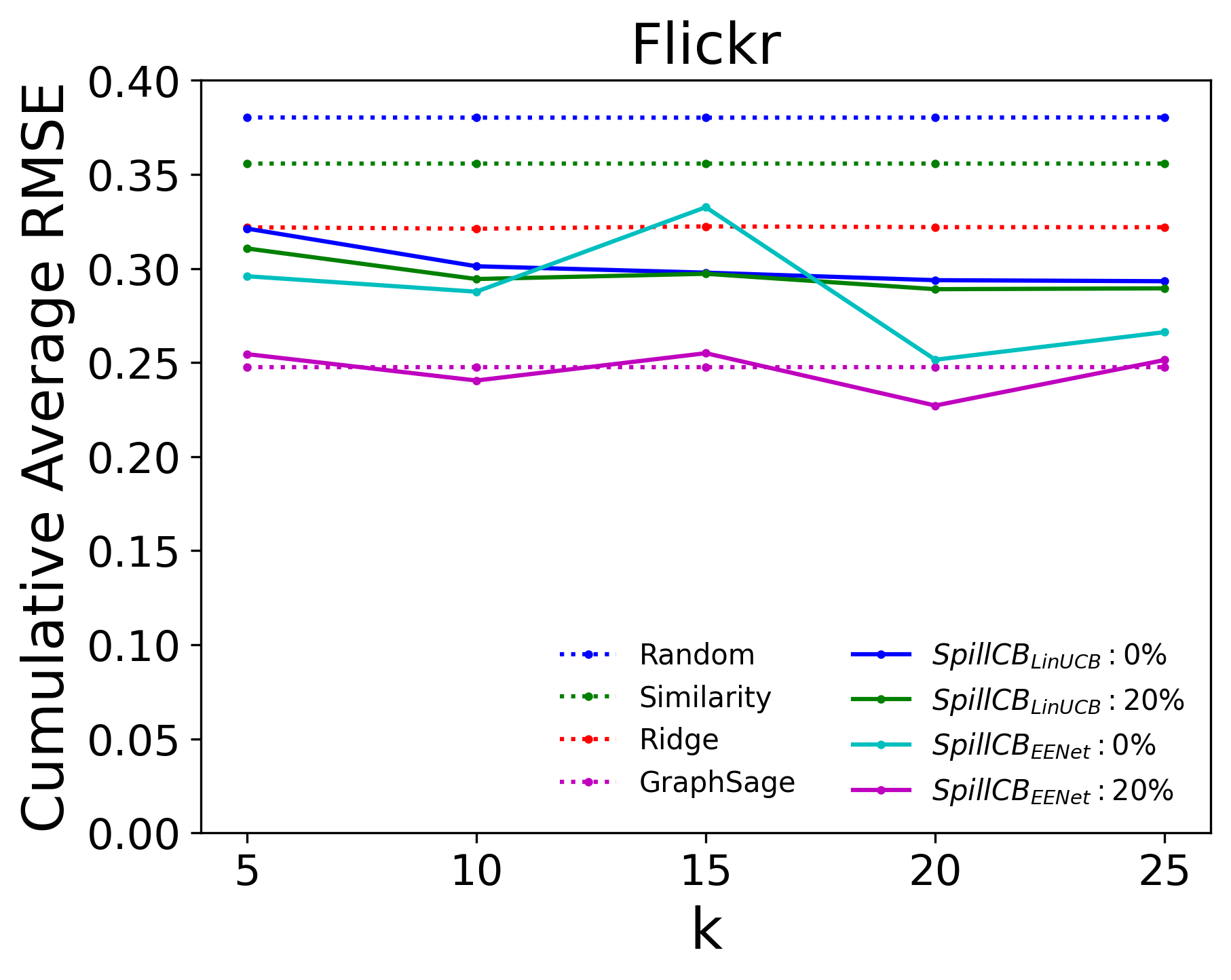}
     \end{subfigure}
     \begin{subfigure}
         \centering \includegraphics[width=0.24\textwidth]{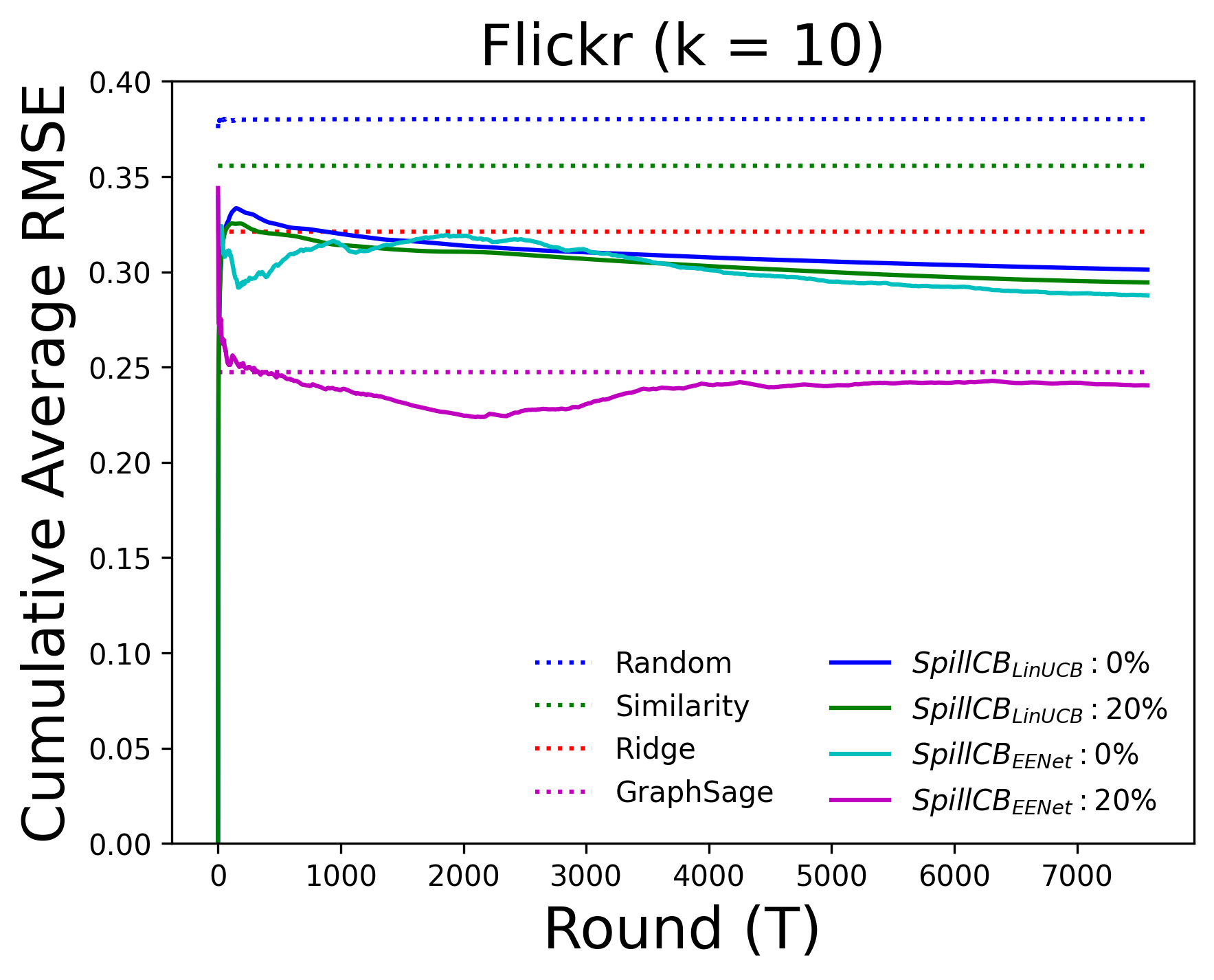}
     \end{subfigure}
     \begin{subfigure}
         \centering \includegraphics[width=0.24\textwidth]{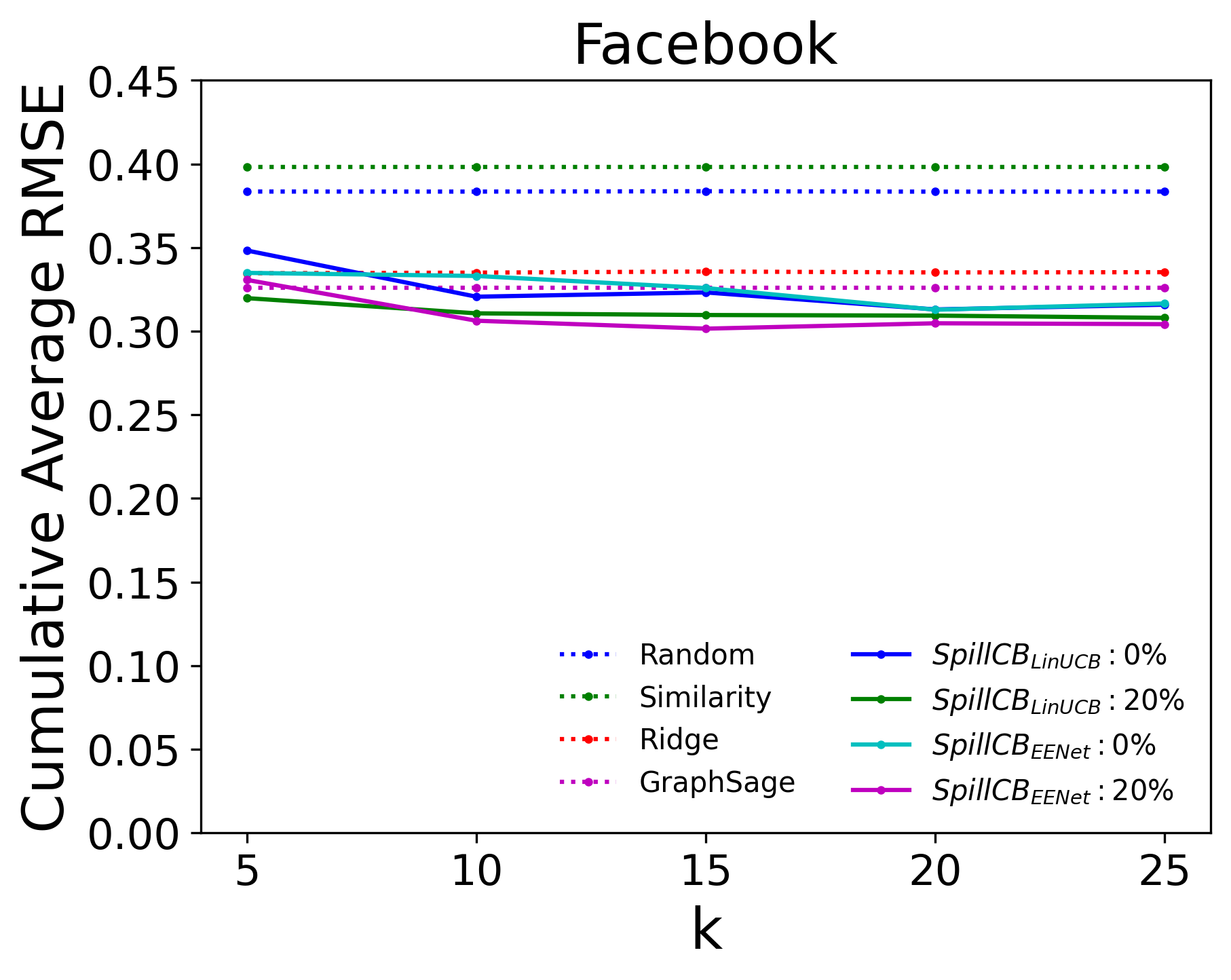}
     \end{subfigure}
     \begin{subfigure}
         \centering \includegraphics[width=0.24\textwidth]{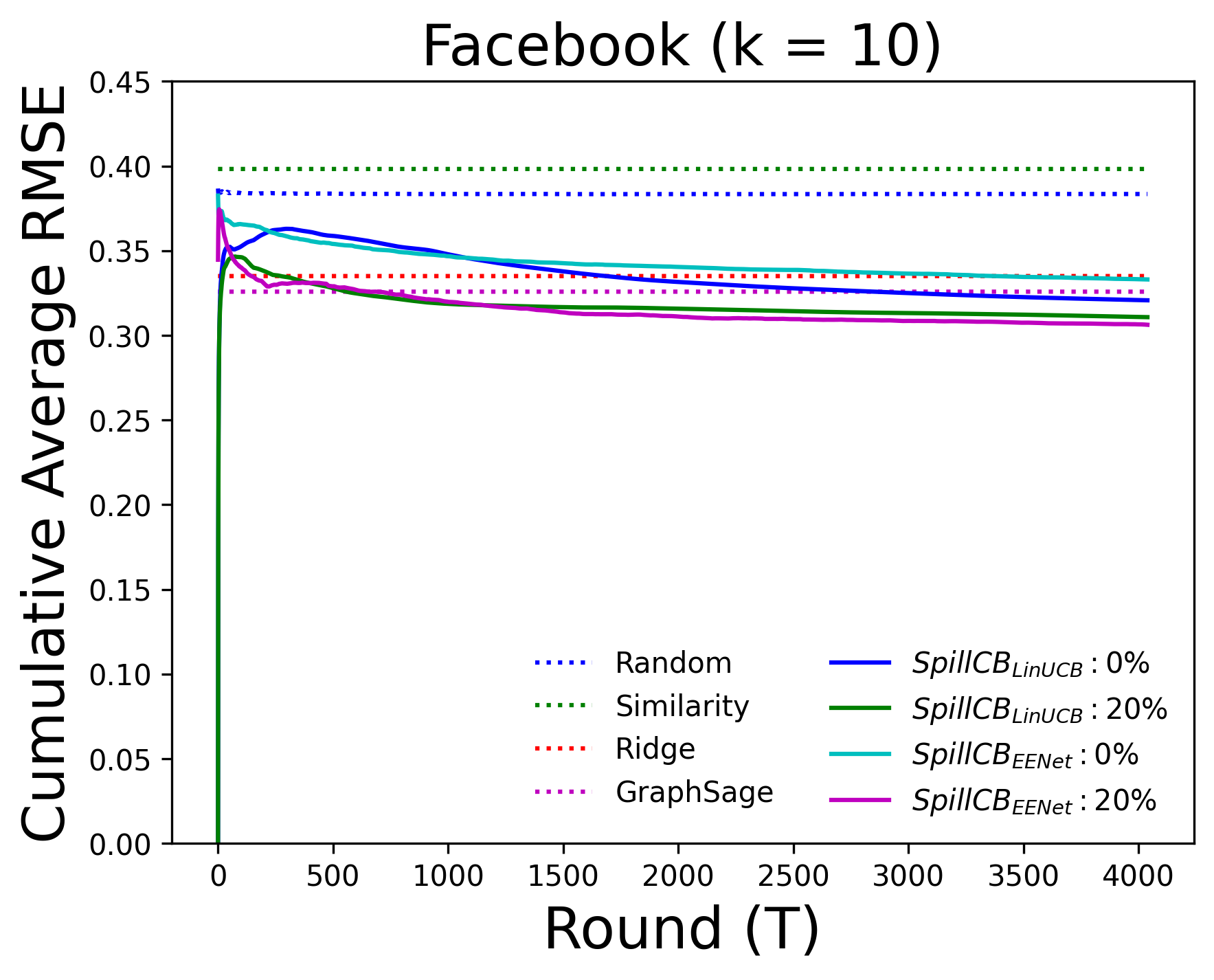}
     \end{subfigure}
     \caption{Comparison of cumulative $\Bacc$ and $RMSE$. Columns $1$ and $3$ show the metrics in the final round for varying k. Columns $2$ and $4$ show changes in the metrics over $T$ rounds for $k = 10$.}
     \label{recall_fixed_k}
     \vspace{-0.52em}
\end{figure*}


\subsection{Evaluation metrics} We evaluate the performance of the $\Lhse$ using the following metrics:
\subsubsection{Recall@$k$} The bandit recall, $\Bacc$ quantifies the $\Lhse$'s ability to correctly identify top-$k$ neighbors. It is defined as the ratio of correctly predicted neighbors belonging to the true top-$k$ neighbors to the total number of true top-$k$ neighbors, i.e., $\Bacc = \sum_{i = 1}^{T}\frac{|N_i \cap N^{*}_{i}|}{|N^{*}_{i}|}$
\subsubsection{Root Mean Squared Error ($RMSE$)} 
To evaluate the accuracy of the learned spillover probabilities, we compute the $RMSE$ on a randomly sampled subset of edges, $\overline{E}\subset E$
from the network. In each round $i$, $RMSE$ is calculated as: $RMSE = \sqrt{\frac{\sum_{e_{xy} \in \overline{E}}{(\hat{e}_{xy}.p - e_{xy}.p)^2}}{|S|}}$ where $\hat{e}_{xy}.p = g(E[R(e_{xy}.X, a_{1})])$ is the estimated spillover probability derived from the estimated rewards $E[R(e_{xy}.X, a_{1})]$
and $g$ is a function to convert the estimated rewards to a probability. The function calculates $z$-score over the set of edges $\overline{E}$ in each round $i$ and then uses a standard normal distribution's cumulative distribution function to convert the $z$-score to a probability.
\subsection{Main algorithms and baselines} We evaluate the performance of our proposed algorithm by comparing it with several baseline approaches:
\subsubsection{Random} This baseline randomly selects $k$ neighbors for recommendation to each system-activated node.
\subsubsection{Similarity-based (Similarity)}This method recommends neighbors based on their proximity to the system-activated node. The top-$k$ neighbors are chosen based on the ascending order of the Euclidean distance between the attribute vectors of the system-activated node and its neighbors. 
\subsubsection{Ridge regression (Ridge)} This method ranks and recommends the top-$k$ neighbors based on link weight prediction scores obtained via ridge regression. All existing edges $E$ are labeled with 1, while $|E|$ non-existing edges are randomly sampled and labeled with 0. Each edge, whether existing or non-existing, is represented by a concatenated attribute vector of its two nodes.
A ridge regression model is trained on these labeled edges to generate link weight prediction scores.
The top-$k$ neighbors of a system-activated node are recommended based on the descending order of their edge scores.
\subsubsection{GraphSage.} 
This approach utilizes node embeddings generated from the GraphSAGE algorithm~\cite{hamilton-neurips-2017}. The approach is similar to the Ridge method, except that the concatenated embeddings are passed through a neural network-based scoring function to generate link weight scores for each edge in the network.
\subsubsection{\textbf{\boldmath SpillCB$_{\textbf{CMAB}}$: z\%.}}
This is our proposed framework, and we consider several variants. We denote the $\Lhse$ as $\Lhse_{LinUCB}$: $z\%$ and $\Lhse_{EENet}$: $z\%$ when the underlying off-the-shelf CMAB algorithm is LinUCB~\cite{li-www-2010} and EENet~\cite{ban-iclr-2022}, respectively. We also vary the amount of exploration $z\%$.
\subsection{Experimental setup} 
Prior to the start of a bandit experiment, we set $v_{i}.y = 0$ for all nodes $v_{i} \in V$, and $e_{ij}.t = 0$ for all edges $e_{ij} \in E$. In each round, we consider the system activates each arrival node and all of its neighbors are inactive. Therefore, our experiments allow re-activation, enabling each node to be activated by multiple neighbors. All experiments are repeated $5$ times, and we report the average results across these repetitions. To compute
$RMSE$ in each round, we sample a fixed subset of edges with size $|S| = 500$. We set $\alpha = 0.5$ for both Ridge and LinUCB methods. 
For EE-Net~\cite{ban-iclr-2022}, we follow their default setup and perform a grid search for the learning rate over $\{0.001, 0.01, 0.1\}$ in their exploitation neural network. The optimal parameters for each dataset are selected from the grid search results. We experiment with  
$z \in \{0, 20\}$ and refer to these configurations as $\Lhse_{CMAB}$: $0\%$
and $\Lhse_{CMAB}$: $20\%$, respectively. We report the corresponding $Recall@k$ and $RMSE$ for $k = 10$ across all rounds in Fig~\ref{recall_fixed_k}. To understand the impact varying $k$, experiments are conducted for $k \in \{5, 10, 15, 20, 25\}$. Results for  $Recall@k$ and $RMSE$ are reported in the final round of the bandit experiments in Figure~\ref{recall_fixed_k}. 
\subsection{Experimental results}
As Figure~\ref{recall_fixed_k} shows, the $Recall@k$ of all methods increases as $k$ increases which is expected. Our proposed methods, $\Lhse_{LinUCB}$ and $\Lhse_{EENet}$, consistently outperform baseline approaches in $Recall@k$ across all values of $k$ on both datasets. A larger $k$ leads to lower 
$RMSE$ because it allows for the exploration of more edges, thereby accelerating the learning process. For instance, the $RMSE$ decreases by $8.72\%$ and $9.31\%$ on the Flickr and Facebook datasets, respectively, as $k$ increases from $5$ to $25$ in $\Lhse_{LinUCB}$: $0\%$. Similarly, $RMSE$ reductions of
$10.04\%$ and $5.47\%$ are observed on the Flickr and Facebook datasets, respectively, for $\Lhse_{EENet}$: $0\%$.

As the number of rounds ($T$) increases, $Recall@10$ improves, while the corresponding $RMSE$ decreases in $\Lhse$ for both datasets. This trend occurs because more edges are explored over rounds, enabling more efficient bandit learning. Moreover, increasing the exploration phase ($z$) results in a reduction in $RMSE$, albeit sometimes at the expense of lower $Recall@10$ in $\Lhse$.  For instance, when the exploration phase increases from $0\%$ to $20\%$, the cumulative average $RMSE$ in the final round decreases by $2.23\%$ for $\Lhse_{LinUCB}$ on the Flickr dataset and by $8.04\%$ for $\Lhse_{EENet}$ on the Facebook dataset. However, this comes with corresponding decreases of $4.2\%$ and $1.73\%$ in $Recall@10$.

These results highlight the trade-offs between the exploration and exploitation phases in our framework and emphasize the robustness of balancing these factors to optimize performance.
\section{Conclusion and Future Work} We present a novel contextual bandit framework, $\Lhse$, designed to model and learn heterogeneous spillover probabilities in networks. Our approach leverages these probabilities to optimize word-of-mouth rewards by recommending top-$k$ neighbors. Our preliminary results look promising. Our next steps include considering more datasets and baselines to further understand the strengths and limitations of the proposed approach.

\section{Acknowledgments}
This research was funded in part by NSF under grant no. 2047899 and DARPA under contract number HR001121C0168.

\bibliography{aaai25}

\end{document}